\documentclass{bmvc2k}

\title{TennisVid2Text: Fine-grained Descriptions for Domain Specific Videos} 

\addauthor{Mohak Sukhwani}{http://researchweb.iiit.ac.in/~mohak.sukhwani/}{1}
\addauthor{C.V. Jawahar}{http://www.iiit.ac.in/~jawahar/}{1}

\addinstitution{
 CVIT\\
 IIIT Hyderabad,\\
 India.\\
 http://cvit.iiit.ac.in
}

\runninghead{Sukhwani, Jawahar}{Descriptions for Domain Specific Videos}

\def\eg{\emph{e.g}\bmvaOneDot}

\def\etal{\emph{et al}\bmvaOneDot}

\usepackage{multirow}
\usepackage{booktabs} 
\usepackage{sidecap}
\usepackage{verbatim}
\usepackage{mathtools}
\begin{document}

\sloppy

\maketitle

\begin{abstract}
Automatically describing 
videos has ever been fascinating. In this work, we 
attempt to describe videos from 
a specific domain -- broadcast videos of lawn tennis 
matches. Given a video shot from a tennis match, we 
intend to generate a textual commentary similar to what a human expert would write on a 
sports website. Unlike many recent works that focus on 
generating short captions, we are interested in generating 
semantically richer descriptions. This demands a detailed low-level analysis of the 
video content, specially the actions and interactions among subjects. 
We address this by limiting our 
domain to the game of lawn tennis. Rich descriptions are generated 
by leveraging a large corpus of 
human created descriptions harvested from Internet.
We evaluate our method on a newly created tennis video data set. 
Extensive analysis demonstrate that our approach 
addresses both semantic correctness as well as 
readability aspects involved in the task.

\end{abstract}

\section{Introduction}
\label{sec:intro}

\begin{figure}[t]
\includegraphics[scale=0.45]{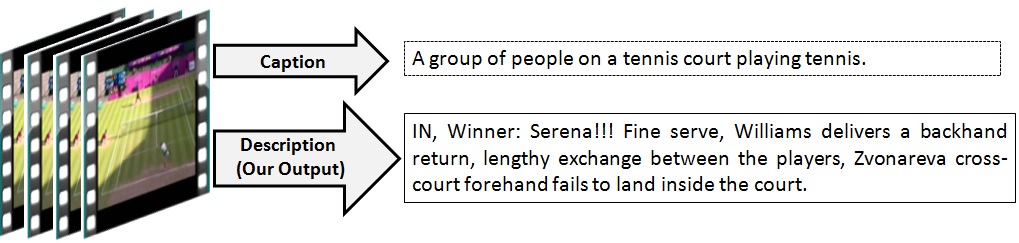}
\caption{For a test video, the {\em caption} generated 
	using an approximation of a state-of- the-art 
	method~\cite{karpathy}, and the {\em description} 
	predicted using our approach. 
	Unlike recent methods that focus on short and 
	generic captions, we aim at detailed and specific 
	descriptions. 
}
\label{fig:capVSdes}
\end{figure}

Annotating visual content with text has 
attracted significant attention in recent years 
~\cite{Sergio,Niveda,xmodal,karpathy,babytalk,BergIm2Txt,Yash1}.  
While the focus has been mostly on images 
~\cite{xmodal,karpathy,babytalk,BergIm2Txt,Yash1}, of late few methods have also been proposed for 
describing videos~\cite{Sergio,Niveda}. 
The descriptions produced by such methods capture the 
video content at certain level of semantics. However, 
richer and more meaningful descriptions may be required 
for such techniques to be useful in real-life applications. 
This becomes challenging in a domain independent 
scenario due to almost innumerable possibilities. We make an attempt towards 
this goal by focusing on a domain specific setting -- lawn tennis videos. 
For such videos, we aim to predict detailed 
(commentary-like) descriptions rather than small captions. 
Figure~\ref{fig:capVSdes} depicts the problem 
of interest and an example result of our method. It even depicts the difference between a caption and 
a description. Rich description generation demands deep 
understanding of visual content and their associations with natural text. 
This makes our problem challenging.   

For the game of tennis, which has a pair of players 
hitting the ball, actions play the central role. 
Here, actions are not just simple verbs like `running', 
`walking', `jumping' etc. as in the early days 
of action recognition but complex and compound 
{\it phrases} like `hits a forehand volley', 
`delivers a backhand return' etc. 
Although learning such activities add to the complexities 
of the task, yet they make our descriptions diverse and vivid. 
To further integrate finer details into descriptions, we 
consider constructs 
that modify the effectiveness of nouns and verbs. Though phrases like 
`hits a nice serve' ,`hits a good 
serve' and `sizzling serve' describe similar 
action, `{\it nice}', `{\it good}' and `{\it sizzling}' 
add to the intensity of that action. We develop a model 
that learns the effectiveness of such phrases, and builds 
upon these to predict florid descriptions.
Empirical evidences demonstrate that our approach predicts 
descriptions that match the videos. 

\section{Related Work} 
Here we briefly discuss some of the works related to 
action recognition, image and video description generation and 
tennis video analysis.\\ 
{\bf Action recognition:} This has been studied in a variety 
of settings~\cite{Schuldt,Efros03}. Initial attempts 
used laboratory videos of few well defined actions. 
In recent years the interest has been shifted to the actions 
captured in the natural settings like movies
~\cite{Laptev,Duchenne}. In all these cases, the actions of interest are 
visually and semantically very different. At times 
when there is low inter-action and high intra-action 
variability in human actions, people have used fine 
grained action classifications to distinguish between 
subtle variations~\cite{fineGrain1}. 
\\ 
{\bf Image and video description:} 
Template based approaches 
~\cite{babytalk,Khan,Lee,Kojima,Sergio,Niveda} 
have been very popular for describing the visual content since the beginning. 
In case of videos, recent works~\cite{Niveda,Sergio} have 
focused on complex task of recognizing 
{\it actor-action-object} relationships rather than 
simple nouns, verbs etc. E.g.,  
Niveda~\etal ~\cite{Niveda} generate descriptions for short videos 
by identifying the best {\sc svo} (\textit{subject-verb-object}) 
triplet. Sergio~\etal ~\cite{Sergio} extend this for out-of domain actions 
(not in training set) by  generating brief sentences 
that sum up the main activity in a video with broader 
verb and object coverage. Barbu~\etal~\cite{Andrei} 
produce rich sequential descriptions of videos using 
detection based tracking and body-posture codebook. 
Template based approaches preserve generality but often 
take away the human aspect of the text. To overcome this, data driven methods 
~\cite{BergIm2Txt,xmodal,Yash1} have been used to predict text for query images. 
Rather than generating a description, these methods retrieve 
the best matching description from a text corpus. These methods often have an 
advantage of producing syntactically better results, 
as retrieved descriptions are human-written and thus well formed. 
\\ 
{\bf Tennis video analysis:} 
Early attempt~\cite{MiyamoriIsk} for 
tennis video analysis and annotation focused on detecting relative 
positions of both players with respect to court 
lines and net to determine action types. Use of 
special set-ups and other audio-video cues to detect players, ball and court lines 
with utmost precision were soon followed~\cite{Pingali,Christmas05asystem}. 
While a few attempts have been made to automate commentary 
generation in simulated settings 
(\eg, for soccer~\cite{Rocco,MIKE}), there has not been much success 
for games in natural settings. Our approach is designed to be effective in real-life 
game play environments and does not assume any 
special/simulated set-up. 
  
\begin{figure}
  \begin{tabular}{cc}
   \includegraphics[width=360pt]{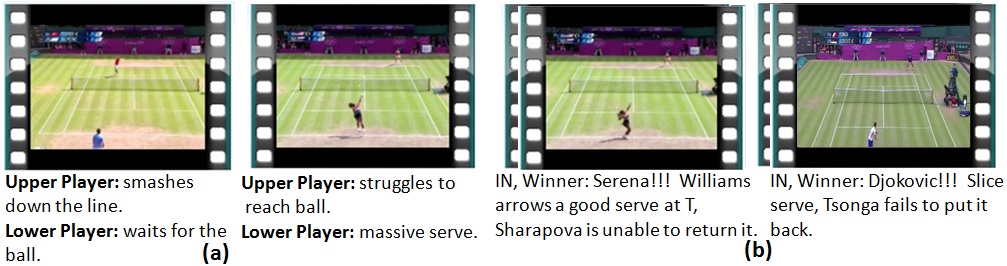}
  \end{tabular} 
\caption{Dataset contents: 
	(a) Annotated-action dataset: short videos 
	aligned with verb phrases. (b) Video commentary dataset: 
	game videos aligned with commentary.}
\label{fig:dataSet}
\end{figure}
\section{Background, Motivation and Dataset Description} 

\begin{table}[b]
\begin{center} 
\begin{tabular}{|l|c|c|}
\hline
\scriptsize{\textbf{Name}} & \scriptsize{\textbf{Contents}} & \scriptsize{\textbf{Role}}\\
\hline
\scriptsize{Annotated-action}  & \scriptsize{$250$ action videos and phrases} & \scriptsize{Classification and Training} \\		
\hline
\scriptsize{Video-commentary}  & \scriptsize{$710$ game videos and commentaries.} & \scriptsize{Testing} \\
\hline
\scriptsize{Tennis Text}	&	\scriptsize{$435K$ commentary lines} & \scriptsize{Dictionary Learning, Evaluation and Retrieval}\\	
\hline
\end{tabular}
\end{center}
\caption{Dataset statistics: 
	Our dataset 
	is a culmination of three standalone datasets. 
	Table describes them in detail, along with 
	the roles they play in the experiments.}
\label{table:tabDataSet}
\end{table}

Lawn tennis is a racquet sport played either individually 
against a single opponent ({\it singles}), or between two 
teams of two players each ({\it doubles}). We restrict 
our attention to singles.  
Videos of such matches have two players -- one in the upper half and the 
other in the lower half of a video frame. 
A complete tennis match is an amalgamation of 
sequence of `tennis-sets', each comprising of a sequence 
of `tennis-games' played with service 
alternating between consecutive sets. A `tennis-point' is the 
smallest sub-division of match that begins with the start of the service and ends 
when a scoring criteria is met. We work at this granularity. 


We use broadcast video recordings for five matches 
from \textit{London Olympics $2012$} for our experiments. 
The videos used are of resolution $640\times360$ at $30$ 
fps. Each video is manually segmented into shots 
corresponding to `tennis-points', and is described with 
a textual commentary obtained  from
~\cite{tennisEarth}. This gives a collection of 
video segments aligned with corresponding commentaries. In total, there 
are $710$ `tennis-points' of average frame length $155$. We refer to this collection as `Video-commentary' dataset.  
This serves as our test dataset and is used for the final evaluation. 
In addition to this, we create an independent 
`Annotated-action' data set comprising $250$ short 
videos (average length of $30$ frames) describing player actions with verb phrases. 
Examples of the verb phrases include \textit{`serves in 
the middle'}, {\it`punches a volley'}, 
{\it`rushes towards net'}, etc.  
In total, we have $76$ action phrases. 
We use this collection to train our action classifiers. 
Figure~\ref{fig:dataSet} shows samples from our dataset. 

As an additional linguistic resource  for creating 
human readable descriptions, 
we crawl tennis commentaries 
(with no corresponding videos). This text corpus 
is built using (human-written) commentary of \textit{$2689$} lawn tennis matches played between 
\textit{$2009$-$14$} from~\cite{tennisEarth}. A typical 
commentary describes the players names, 
prominent shots and the winner of the game. 
We refer this collection as `Tennis-text'. 
Table~\ref{table:tabDataSet} summarizes 
the three datasets. 
Note that all the datasets are independent, with 
no overlap among them. 

\begin{SCfigure}
\includegraphics[width=210pt, height=135pt]{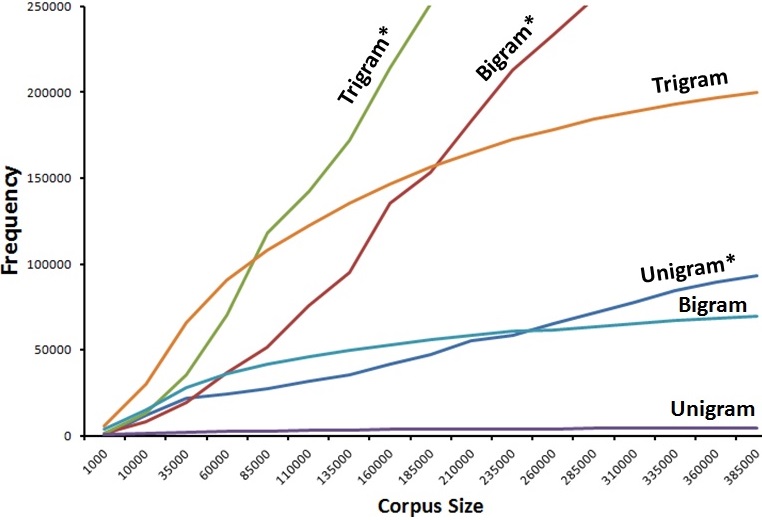}
\caption{Trend illustrating saturation of word-level 
	trigram, bigram and unigram counts 
	in domain specific settings. Emphasized(*) labels indicate 
	corpus of unrestricted 
	tennis text (blogs/news) and remaining labels indicate tennis commentary 
	corpus. Bigram* and Trigram* frequencies 
	are scaled down by a factor of $10$ to show the comparison.}
\label{fig:corpusComp}
\end{SCfigure}

We seek to analyse how focusing on a specific domain 
confines the output space. We compute the count of 
unique (word-level) unigrams, bigrams and trigrams in 
tennis commentaries. Each commentary sentence in 
the Tennis-text corpus is processed individually using 
standard Natural Language ToolKit ({\sc nltk}) 
library, and word-level n-gram frequencies of corpus are computed. The `frequency' (count)
trends of unigrams, bigrams and trigrams 
plotted over `corpus size' (number of lines in corpus) are depicted 
in Figure~\ref{fig:corpusComp}.  
We compare these with the corresponding 
frequencies in unrestricted tennis text mined from on-line tennis news, blogs, etc. 
(denoted by `*' in the figure). 
It can be observed that in case of tennis 
commentary, the frequency of each n-gram 
saturates well within a small corpus as 
compared to corresponding frequencies of 
unrestricted text. The frequency plots reveal that the vocabulary 
specific to tennis {\em commentary} is indeed small, 
and sentences are often very similar. Hence, in a domain specific environment, 
we can create rich descriptions even with a limited 
corpus size. 


\section{Approach}
\label{sec:Approach} 

Our goal is to automatically describe video segments 
of {\it tennis points} in the Video-commentary dataset. 
We begin by learning phrase classifiers using Annotated-action dataset. 
Given a test video, we predict 
a set of action/verb phrases individually for each frame 
using the features computed from its neighbourhood. 
Since this sequence of labels could be noisy, these 
are smoothed by introducing priors in an energy 
minimization framework. The identified phrases along with 
additional meta-data (such as player details) are used 
to find the best matching description from the Tennis-text dataset. 
Major activities during any tennis game take place 
on each side of the net, we analyse videos by 
dividing them across the net in the subsequent 
sections (referred as `upper' and `lower' video/frame). 
In almost all tennis broadcast videos, this 
net is around the center of a frame, and thus can 
be easily approximated. 

\subsection{Court Detection, Player Detection and Player Recognition} 
\label{sec:CourtPlayerDetection} 

\begin{figure}[t]
\begin{tabular}{ccccc}
\bmvaHangBox{\includegraphics[width=75pt, height=70pt]{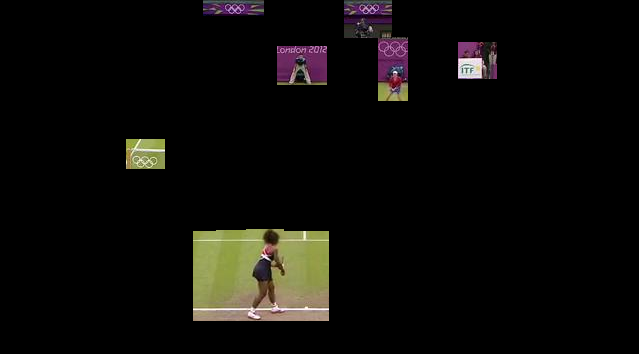}}
\bmvaHangBox{\includegraphics[width=60pt, height=70pt]{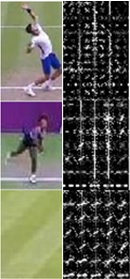}}
\bmvaHangBox{\includegraphics[width=65pt, height=70pt]{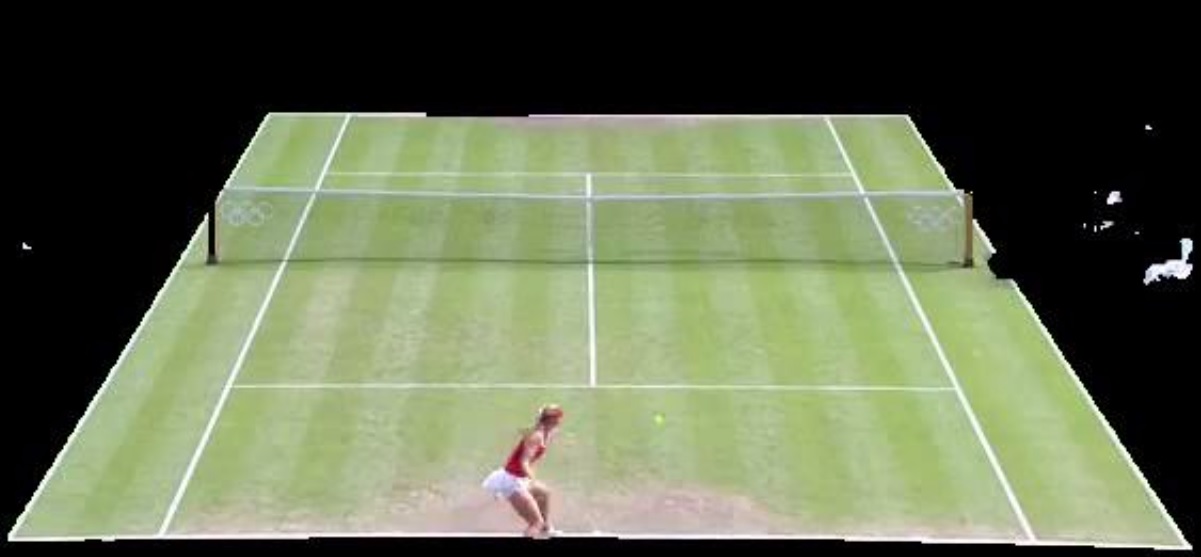}}
\bmvaHangBox{\includegraphics[width=75pt, height=70pt]{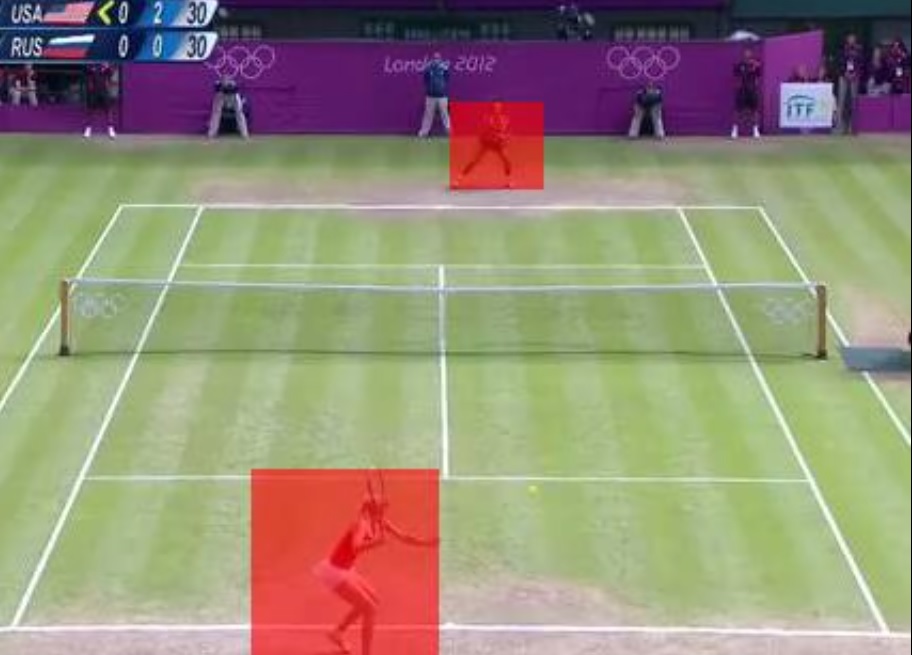}}
\bmvaHangBox{\includegraphics[width=75pt, height=70pt]{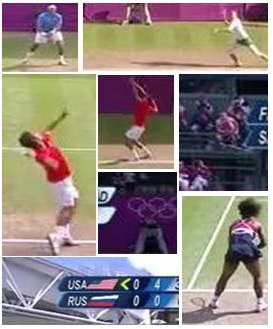}}\\
(a) \hspace{1.75cm} (b) \hspace{1.75cm} (c) \hspace{1.75cm} (d) \hspace{1.75cm} (e)\\
\end{tabular}
\caption{Retrieving player details: (a) Extracted 
	foreground regions. (b) Visualization of 
	\textsc{hog} features for players and non-player 
	regions. (c) Court detection. (d) Player Detection. 
	(d) Some examples of successful and failed player 
	detections. 
}
\label{fig:playerDetect}
\end{figure}

We begin by identifying/tracking both players 
on the tennis court. The playing court in lawn tennis has a set of prominent 
straight white lines, each of which adds meaning to the game 
in a unique way. We detect these lines using the 
Hough transform and consider only the most prominent lines 
(by keeping a threshold on length). 
A bounding-box is created encompassing the set of 
identified lines, which is then  
considered as foreground seed 
for the GrabCut algorithm~\cite{grabCut}. This 
in turn returns the playing 
field as shown in Figure~\ref{fig:playerDetect}(c).     

In a tennis broadcast video background is (nearly) static 
after the serve begins, and remains the same till a player wins a point. 
Based on this, the candidate regions for players are segmented 
through background subtraction 
using thresholding and connected component analysis. 
Each candidate foreground region thus obtained is 
represented using {\sc hog} descriptor~\cite{hog}. 
To prune away false detections (i.e., non-player regions), multi-class 
{\sc svm} with {\sc rbf}-kernel is employed distinguishing 
between `upper-player', `lower-player' and `no player' 
regions. Figure~\ref{fig:playerDetect}(b) visualizes {\sc hog} features for few 
examples. The detected windows thus obtained are used to recognize players.
In any particular tournament (and in general) 
players often wear similar colored jerseys and 
depict unique stance during game play, we use these cues to recognize them.
We perceive both color and stance information using \textsc{cedd}~\cite{CEDD} descriptor.
This descriptor captures both colour and edge (for stance) information 
of each detected candidate player region. We use Tanimoto distance~\cite{CEDD} to build our classifier. 
Classifier scores averaged over initial ten frames are used to 
recognize both the players.
Figure~\ref{fig:playerDetect}(d) 
highlights the players detected in a frame, and 
Figure~\ref{fig:playerDetect}(e) shows 
some true and false detections. 

\subsection{Learning Action Phrases}
 
\label{sec:LearningActionPhrases} 
\begin{figure}[t]
 \includegraphics[width=370pt,height=40pt]{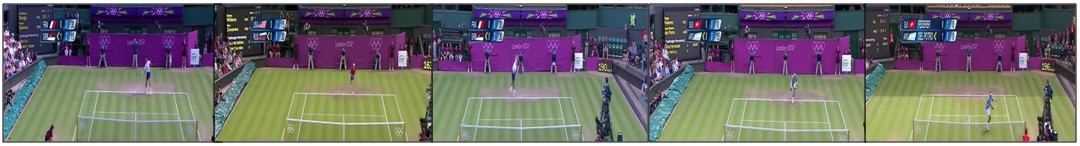}
 \includegraphics[width=370pt,height=40pt]{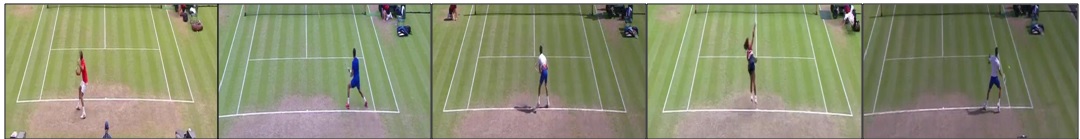}
\caption{Example frames 
	depicting varied actions. Upper frames are 
	shown at the top,  
	and lower frames at the bottom. 
	Here, upper and lower frames do 
	not correspond to same video. 
}
\label{fig:finalRes}
\end{figure}

We learn phrase classifiers using 
`Annotated-action' dataset. 
For representation, we use descriptors as described in ~\cite{Wang2011}, 
and extract dense trajectory features 
over space-time volumes (using default parameters). 
For each trajectory, we use Trajetory, 
\textsc{hog}, \textsc{hof} 
(histograms of optical flow)~\cite{hof} and 
\textsc{mbh} (motion boundary histogram)~\cite{mbh} 
descriptors.  
While \textsc{hog} captures static appearance information, 
\textsc{hof} and \textsc{mbh} measure motion information 
based on optical flow. The dimensions of each of 
these descriptors 
are: $30$ for Trajectory, $96$ for \textsc{hog}, $108$ 
for \textsc{hof} and $192$ for \textsc{mbh}. 
For each descriptor, bag-of-words ({\sc bow})
representation is adopted (with vocabulary size $2000$). We take square root 
of each element in a feature vector before computing the 
codebook (similar to 
RootSIFT~\cite{Arandjelovic12}). The final representation is concatenation 
of {\sc bow} histograms of all the descriptors. 
Using this, a 1-vs-rest  
{\sc svm} classifier (with {\sc rbf} kernel and 
$\chi^2$ distance) is learned for each phrase. 
In all, we have $76$ verb phrases - $39$ for upper and $37$ for lower 
player. Figure~\ref{fig:finalRes} illustrates some examples of player actions. 

\subsection{Verb Phrase Prediction and Temporal Smoothing} 

Given a (test) video, we recognize verb phrase 
for each 
frame by extracting features from neighbouring frames 
using sliding window
(neighbourhood of size $30$ frames). 
Since this typically 
results into multiple firings, non-maximal suppression 
({\sc nms}) is applied. 
This removes low-scored responses that are 
in the neighbourhood of 
responses with locally maximal confidence scores. Once we get
potential phrases for all windows 
along with their scores, we remove the independence 
assumption and smooth the predictions using an energy 
minimization framework. For this, a 
Markov Random Field ({\sc mrf}) based model is used 
which captures dependencies among nearby phrases. 
We add one node for each window sequentially 
from left to right and connect these by edges. 
Each node takes a label from the set of 
action phrases. 
The energy function for nodes $\nu$, neighbourhood 
$\cal N$ and labels $\cal L$ is:
\begin{equation} 
E = \sum_{p \in \nu } D_p(f_p) + 
\sum_{p,q \in \cal N} V_{pq}(f_p,f_q) 
\end{equation}

Here, $D_p(f_p)$ denotes \textit{unary phrase 
selection cost}. This is set to $1-p(l_j|x_p)$, 
where $p(l_j|x_p)$ is the \textsc{svm} score 
of the phrase label $l_j$ for node $x_p$, 
normalized using the Platt's method~\cite{platt}. 
The term $V_{pq}(f_p,f_q)$ denotes 
\textit{pairwise phrase cohesion cost} 
associated with two neighbouring 
nodes $x_i$ and $x_j$ taking some phrase label. 
For each pair of phrases, this is determined by their 
probability of occurring together in the game play, 
and is computed using frame-wise transition probability. 
In our case, since there are two players, and each player's 
shot depends on the other player's shot and his 
own previous shot, we consider four probability scores: 
$p(l_{iP_1},l_{jP_1})$, $p(l_{iP_1},l_{jP_2})$, 
$p(l_{iP_2},l_{jP_2})$ and $p(l_{iP_2},l_{jP_1})$. 
Here, $p(l_{iP_1},l_{jP_2})$ refers to the probability of phrase 
$l_i$ of $player 1$ and $l_j$ of $player 2$ occurring 
together during game play. We compute pairwise cost, $1-p$, 
for each of the four probabilities and
solve the minimization problem using 
a loopy belief propagation ({\sc bp}) algorithm~\cite{LoopyBP}.

\subsection{Description Prediction} 
\label{sec:DescriptionPrediction}

Let, $W=\{w_1,w_2,\ldots, w_n\}$ 
be set of unique words present in the
group of phrases along with player names, and 
$S=\{s_1,s_2,\ldots, s_m\}$ be the set of all 
the sentences in the Tennis-text corpus. Here, each 
sentence is a separate and full commentary description. 
We formulate the task of predicting the final 
description as an optimization problem of selecting the 
best commentary from $S$ that covers as many 
words as possible. 
Let, $x_i$ be a variable which is $1$ if sentence 
$s_i$ is selected, and $0$ otherwise. Similarly, 
let $a_{ij}$ be a variable which is $1$ if sentence 
$s_i$ contains word $w_j$, and $0$ otherwise. 
A word $w_j$ is said to be covered if it is 
present in the selected sentence ($\sum_{i}a_{ij}x_i=1$). Hence, our 
objective is to find a sentence that covers as many 
words as possible:
\begin{equation} 
  \max \smashoperator{\sum_{i \in \{1,2,\ldots,m\},j \in \{1,2,\ldots,n\}}} 
  a_{ij}x_i, 
  \;\;\;\;\;\;
  s.t.  \sum_{i=1}^{m}x_i =1, 
  \;\;
  \forall a_{ij}, x_i \in \{0,1\}
\end{equation}

\begin{figure}[t]
 \includegraphics[width=370pt]{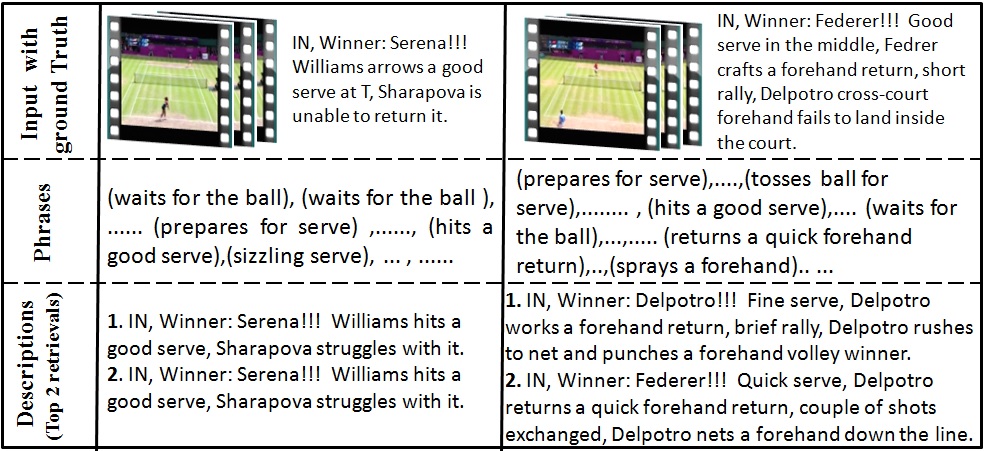}
\caption{Illustration of our approach. Input sequence 
	of videos is first translated into a set 
	of phrases, which are then used to produce 
	the final description. 
}
\label{fig:illustration}
\end{figure}

In the above formulation, doing na\"ive lexical matching 
can be inaccurate as it would consider just the presence/absence of words, and 
fail to capture the overall semantics of the text. 
To address this, we adopt
Latent Semantic Indexing (\textsc{lsi})~\cite{lsi}, 
and use statistically derived conceptual 
indices rather than individual words. \textsc{lsi} 
assumes an underlying structure in word 
usage that is partially obscured by variability in 
word choice. It projects derived phrases and corpus 
sentences into a lower dimensional subspace, 
and addresses the problems of synonymy ({similar meaning words}). 
Figure~\ref{fig:illustration} illustrates the steps involved 
in our method by taking two examples. The 
verb phrase prediction and smoothing 
steps provide a set of relevant phrases. Number of such phrases depend on the size of the 
(test) video. This is evident from the second example (right), which 
is of longer duration and thus has 
more phrases predictions.  
These phrases are used
to select the best matching commentary from 
the Tennis-text corpus. Since similar events are described by
identical descriptions in text corpus, there could be 
instances where the retrieved descriptions are 
same -- first example (left) in Figure~\ref{fig:illustration}. 

\section{Experiments and Results} \label{sec:expSec}
\subsection{Experimental Details} \label{sec:ExperimentalDetails} 
{\bf (1) Creating Textual Dictionary:}
We create textual dictionary 
using commentary descriptions from our Tennis-text corpus. 
The text is processed using
standard \textsc{nltk} modules. 
This involves tokenizing, filtering out common stopwords and 
stemming. The dictionary thus obtained is used to 
compute tf-idf based feature representation of the commentary text.
\\
{\bf (2) Player Detection and Recognition:}
To learn multi-class \textsc{svm} for differentiating 
between player and non-player regions, 
we use manually annotated examples. The bounding boxes 
(separate for `upper-player' and `lower-player') are identified
from random video frames of `Annotated action' dataset. In all we had $2432$ `lower-player', 
$2421$ `upper-player' and $13050$ `no player' windows. Similarly for player recognition, we learn classifiers 
using {\sc cedd} features over `lower-player' and `upper-player' 
windows. Each window in this case is labelled as one of the eight unique players 
in our training set. The player recognition classifier is run 
over the candidate player-regions proposed 
by the previous module. 
\\
{\bf (3) Evaluation Criteria:}
We conduct both automatic as well as human evaluations 
to validate our approach. 
For automatic evaluation, we consider 
\textsc{bleu} score that has been popularly used by 
several other relevant works such as
~\cite{babytalk,BergIm2Txt,Sergio,Niveda,Yash1,karpathy}. 
It measures n-gram agreement (precision) 
between test and reference text, with higher 
score signifying better performance. For each test video, the scores are averaged over the 
top five retrieved descriptions, by 
matching them with the ground truth commentary. As part of human evaluation, we collected judgements from 
twenty human evaluators with ample tennis exposure. Videos
were presented in form $15$ sets comprising of $6$ videos 
each. Every evaluator was randomly presented (atleast) two sets 
and asked to rate both linguistic structure as well as semantics of a predicted 
description/commentary. The rating was done on a subjective scale of 
\{`Perfect', `Good', `OK', `Poor', `Flawed'\}. 
These were converted on a likert scale 
of $1$-$5$, with $5$ being `Perfect' and $1$ being 
`Flawed'. We report average for both scores. 

\begin{table} 
\begin{tabular}{c||c} 
\begin{tabular}{|ccccccc|}
\hline
\multicolumn{1}{c}{\scriptsize{\bf Corp$\#$}} 
& \scriptsize{\bf Vocab$\#$} & 
\scriptsize{\bf B-1} & \scriptsize{\bf B-2} 
& \scriptsize{\bf B-3} & \scriptsize{\bf B-4} 
\\
\hline
\multicolumn{1}{c}{\scriptsize{100}} 
& \scriptsize{$85$} & 
\scriptsize{$0.379$} & \scriptsize{$0.235$} & 
\scriptsize{$0.154$} & \scriptsize{$0.095$} 
\\
\multicolumn{1}{c}{\scriptsize{500}} 
& \scriptsize{$118$} & 
\scriptsize{$0.428$} & \scriptsize{$0.251$} & 
\scriptsize{$0.168$} & \scriptsize{$0.107$} 
\\
\multicolumn{1}{c}{\scriptsize{5K}} 
& \scriptsize{$128$} & 
\scriptsize{$0.458$} & \scriptsize{$0.265$} & 
\scriptsize{$0.178$} & \scriptsize{$0.111$} 
\\
\multicolumn{1}{c}{\scriptsize{30K}} 
& \scriptsize{$140$} & 
\scriptsize{$0.460$} & \scriptsize{$0.277$} & 
\scriptsize{$0.182$} & \scriptsize{$0.113$} 
\\
\multicolumn{1}{c}{\scriptsize{50K}} 
& \scriptsize{$144$} & 
\scriptsize{$0.461$} & \scriptsize{$0.276$} & 
\scriptsize{$0.183$} & \scriptsize{$0.114$} 
\\
\hline
\end{tabular} 
& 
\begin{tabular}{lcccc}
\hline
\scriptsize{\bf Method} & 
\scriptsize{\bf B-1} & \scriptsize{\bf B-2} & 
\scriptsize{\bf B-3} & \scriptsize{\bf B-4} 
\\
\hline
\scriptsize{Guadarrama~\cite{Sergio}} & 
\scriptsize{$0.119$} & \scriptsize{$0.021$} & 
\scriptsize{$0.009$} & \scriptsize{$0.002$} 
\\
\scriptsize{Karpathy~\cite{karpathy}} & 
\scriptsize{$0.135$} & \scriptsize{$0.009$} & 
\scriptsize{$0.001$} & \scriptsize{$0.001$} 
\\
\scriptsize{Rasiwasia~\cite{xmodal}} & 
\scriptsize{$0.409$} & \scriptsize{$0.222$} & 
\scriptsize{$0.132$} & \scriptsize{$0.070$} 
\\
\scriptsize{Verma~\cite{Yash1}} & 
\scriptsize{$0.422$} & \scriptsize{$0.233$} & 
\scriptsize{$0.142$} & \scriptsize{$0.075$} 
\\
\hline 
\scriptsize{This work} & 
\scriptsize{$0.461$} & \scriptsize{$0.276$} & 
\scriptsize{$0.183$} & \scriptsize{$0.114$} 
\\
\hline 
\end{tabular}
\end{tabular}
\\
\caption{{\bf Left:} Variation in {\sc bleu} score 
	with corpus size. 
	{\bf Right:} Performance comparison with 
	previous methods using the best performing 
	dictionary. `Corpus\#' denotes the number of commentary lines,
	`Vocab\#' denotes the dimensionality of the textual 
	vocabulary/dictionary and `B-n' means n-gram {\sc bleu} score. 
}
\label{tab:corpusCompare}
\end{table} 

\subsection{Comparison Baselines} 
\label{sec:Baselines} 

The proposed system is benchmarked against 
state-of-the-art methods from 
two streams that are popular in predicting 
textual descriptions for visual data: 
description/caption generation 
and cross-modal description retrieval. 
\\ 
{\bf (1) Description generation:} Since the approaches 
in this domain are either too generic~\cite{Sergio,Niveda,karpathy}, 
or designed for images~\cite{karpathy}, 
we evaluate by adapting them to our setting. 
In both~\cite{Sergio,Niveda}, a template-based 
caption is generated by considering a triplet of form (\textsc{svo}). 
To compare with this setting, we align the best predicted verb phrase into a 
template `player1 $-$ verbPhrase1 $,$ player2 $-$ verbPhrase2'. 
Since a verb phrase is a combination of an action verb and an 
object, this template resembles the `\textsc{svo}' selection 
of~\cite{Sergio,Niveda}. 
To compare with~\cite{karpathy}, we use the publicly 
available pretrained model and generate 
captions for key frames in a video. Since 
this is a generic approach, the captions generated 
are nearly similar, and the set of distinct captions 
is far less than the total number of key-frames. 
To associate a caption with a video, we pick 
the one with the highest frequency. 
\\ 
{\bf (2) Cross-modal description retrieval:} 
Cross-modal retrieval approaches~\cite{xmodal,Yash1} 
perform retrieval by matching samples from 
the input modality with those in the output modality 
(both of which are represented as feature vectors). 
While comparing with~\cite{Yash1}, we consider the best 
performing variant of their approach; i.e., the one that 
uses projected features with Euclidean distance as loss function. 
Note that our approach is also based on retrieving a description; however, 
it makes explicit use of low-level visual and textual cues 
unlike cross-modal retrieval approaches. This difference 
is also evident from the experimental results, where 
our approach is shown to retrieve better descriptions than~\cite{xmodal,Yash1}. 

\subsection{Results} 
\label{sec:Results} 

\begin{figure}
\includegraphics[scale=0.48]{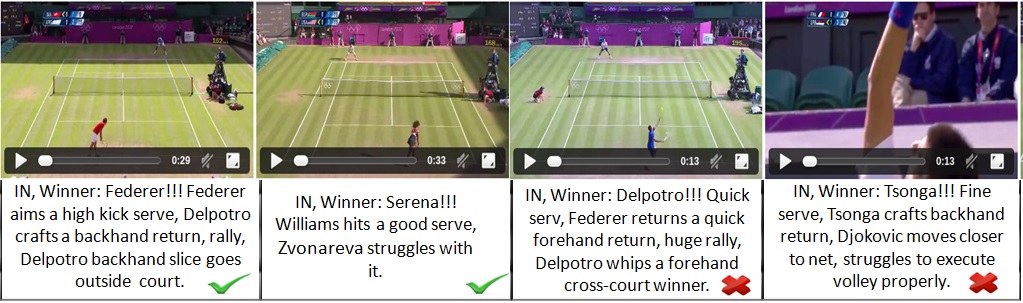}
\caption{Success and failure cases: Example videos 
	along with their descriptions. The `ticked' descriptions match with the ground 
	truth, while the `crossed' ones do not. 
}
\label{fig:tenRes}
\end{figure}

To validate the utility of our design choices, we discuss performance of 
various modules in this section. More details 
on these are provided in the 
supplementary file.

\subsubsection{Performance of Individual Modules} 

\textbf{(1) Verb Phrase Recognition:} 
Compared to a straightforward use of 
state-of-the-art technique~\cite{Wang2011} to 
recognize verbs, our verb phrase recognition module 
performs better by around $15\%$. 
This suggests the utility of harnessing relevant 
cues (dividing a frame into two halves) 
while working in a domain specific environment.\\
\textbf{(2) Smoothing Vs. No Smoothing of Verb Phrase Predictions:} 
In practice, using {\sc mrf} for smoothing phrase predictions 
improved average \textsc{bleu} score from $0.204$ to $0.235$. 
\\
\textbf{(3) LSI-based Matching Vs. Lexical Matching:} 
Employing {\sc lsi} technique while 
matching predicted phrases with descriptions achieves 
an average \textsc{bleu} score of $0.235$, whereas 
na\"ive lexical matching achieves $0.198$. 

\subsubsection{Video Description Performance} 

Table~\ref{tab:corpusCompare} (left) demonstrates the
effect of variations in corpus size on {\sc bleu} 
scores.
It can be observed 
that the scores saturate soon, which validates our 
initial premise that in domain specific settings, rich descriptions 
can be produced even with small corpus size. 
In Table~\ref{tab:corpusCompare} (right), we compare our 
performance with
some of the recent methods. 
Here we observe that caption generation based 
approaches~\cite{Sergio,karpathy} achieve very low 
{\sc bleu} score.~\footnote{Recall that~\cite{Sergio,karpathy} work for 
generic videos and images, we approximate them 
(Section~\ref{sec:Baselines}) for comparisons.} 
This attributes to their generic 
nature, and their current inability to 
produce detailed descriptions. On the other 
hand, cross-modal retrieval approaches~\cite{xmodal,Yash1} 
perform much better than~\cite{Sergio,karpathy}. 
Compared to all the competing methods, our approach 
consistently provides better performance. 
The performance improvements increase as we move 
towards higher n-grams, with an improvement of 
around $50\%$ over~\cite{Yash1} for $4$-gram. 
These results confirm the efficacy of our approach 
in retrieving descriptions that match the semantics 
of the data much better than cross-modal 
retrieval approaches. 
In human evaluation, we achieve an average score 
of $3.21$ for semantics, and $3.9$ for structure of 
the predicted descriptions. The scores reported are on scale of $1$-$5$.
Figure~\ref{fig:tenRes} depicts some success and failure 
examples (success means the topmost predicted description matches the ground-truth description).

\section{Conclusion}
 
We have introduced a novel method for predicting 
commentary-like descriptions for lawn tennis videos. 
Our approach demonstrates the utility of the simultaneous use of vision,
language and machine learning techniques in a domain specific 
environment to produce semantically rich and 
human-like descriptions. 
The proposed method is fairly generic and 
can be adopted 
to similar situations where activities are 
in a limited context and the linguistic diversity is confined, 
however the output description can be semantically rich. 
Applications of our solution could range 
from content based retrieval to real life tennis coaching. 

\bibliography{egbib}
\end{document}